# AUTOENCODER INSPIRED UNSUPERVISED FEATURE SELECTION

*Kai Han, Yunhe Wang, Chao Zhang*, Chao Li, Chao Xu*

Key Laboratory of Machine Perception (MOE) and Cooperative Medianet Innovation Center,
School of EECS, Peking University, Beijing 100871, P.R. China.
`hankai@pku.edu.cn, wangyunhe@pku.edu.cn, c.zhang@pku.edu.cn,`
`li.chao@pku.edu.cn, xuchao@cis.pku.edu.cn`

## ABSTRACT

High-dimensional data in many areas such as computer vision and machine learning tasks brings in computational and analytical difficulty. Feature selection which selects a subset from observed features is a widely used approach for improving performance and effectiveness of machine learning models with high-dimensional data. In this paper, we propose a novel AutoEncoder Feature Selector (AEFS) for unsupervised feature selection which combines autoencoder regression and group lasso tasks. Compared to traditional feature selection methods, AEFS can select the most important features by excavating both linear and nonlinear information among features, which is more flexible than the conventional self-representation method for unsupervised feature selection with only linear assumptions. Experimental results on benchmark dataset show that the proposed method is superior to the state-of-the-art method.

## 1. INTRODUCTION

With the development of the big data technology, we have been given more and more high-dimensional data, which really boost the performance of machine learning models. However, there are considerable noisy and useless features often collected or generated by different sensors and methods, which also occupy a lot of computational resources. Therefore, feature selection acts a crucial role in the framework of machine learning which removes nonsense features and preserves a small subset of features to reduce computational complexity. Wherein, unsupervised feature selection method is much more essential than supervised approaches, since sample labels are often unknown and labelling samples is both time-consuming and finance-consuming in real world applications.

*Corresponding author. This research is partially supported by National Natural Science Foundation (NSF) of China (grant no.61671027) and National Basic Research Program of China (973 Program) (grant no.2015CB352303). We also thank supports of NSFC 61375026 and 2015BAF15B00.

According to different assumptions and strategies, existing feature selection methods can be divided into filter, wrapper, and embedded based methods [1, 2]. Wherein, the embedded method is a research hotspot currently. Compared with filter and wrapper based methods which regard feature selection process and training process as two separate parts, embedded based methods combine variable selection in the training process. Thus, embedded methods have lots of advantages, *e.g.*, being more efficient, interacting with the learning algorithm and saving plentiful time for model training.

However, most of traditional embedded based methods such as the famous LASSO [3] method can only explore the linear relationship among features, which ignore the nonlinear relationship among features. In order to excavate nonlinear information, Kernel based feature selection methods [4, 5, 6] were proposed learning nonlinear representation, but the representation is limited by the fixed kernel [7], and the choice of the optimal kernel or combination of kernels is difficult. In this paper, we propose to use a neural network to learn flexible nonlinear relationships among features, which can learn arbitrary transforms and is able to deal with a variety of tasks (e.g., visual recognition [8, 9] and image segmentation [10]).

In order to explore a more effective unsupervised feature selection method, we propose to use an autoencoder network for selecting features with high representability, which is a widely used neural network for unsupervised learning of efficient codings [11] or supervised dimensionality reduction [12], *etc*. Since the redundant features can be represented by linear or nonlinear combinations of other useful features, the autoencoder network can squeeze input features into a low-dimensional space and represent original features by exploiting these low-dimensional data. Therefore, features with less effect on the low-dimensional data (*i.e.*, hidden units) could be recognized as redundancy, which can be removed by a group sparsity regularization. Experiments conducted on benchmark datasets verify the effectiveness of the proposed method over other methods.

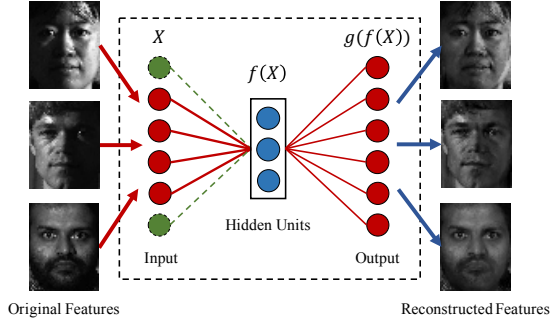

**Fig. 1**. The diagram of the proposed method. An autoencoder network is utilized for excavating useful features (red units and lines) and discarding redundant features (green units and lines) by preserving the representability of original features.

## 2. AUTOENCODER FEATURE SELECTOR

**Preliminaries.** Given the unlabeled sample matrix $X = [\boldsymbol{x}_1, ..., \boldsymbol{x}_m]^T \in \mathbb{R}^{m \times d}$, where $m$ is the number of unlabeled samples, and $d$ is dimensionality of the sample, *i.e.*, the number of features, the task of unsupervised feature selection aims to select $s$ ($s \leq d$) most discriminative and informative features from $X$ with the unlabeled data.

The autoencoder [13] is a special feedforward neural network which receives a set of features and outputs them after applying a serious transforms. We use a simple autoencoder network with two fully connected layers as suggested in [12]. The typical autoencoder with a $h$-dimension hidden layer consists of two components: an encoder function $f(X) = \sigma_1(XW^{(1)})$ and a decoder that produces a reconstruction $\hat{X} = g(f(x)) = \sigma_2(f(X)W^{(2)})$, where $\sigma_1, \sigma_2$ are activation functions (which can be linear or nonlinear ones such as sigmoid, ReLU, tanh, *etc.*) of the hidden layer and the output layer, respectively, $\Theta = \{W^{(1)}, W^{(2)}\}$ are weight parameters and $W_{ij}^{(l)}$ denotes the parameter of the connection between the $i$-th neuron in the $l$-th layer and the $j$-th neuron in the $(l+1)$-th layer. The overall function of the autoencoder could be represented as $g(f(X))$.

In the learning process, autoencdoer is simply described as minimizing a loss function, which is set as the following least square loss in the proposed method for seeking the self-representation:

$$\mathcal{J}(\Theta) = \frac{1}{2m}\|X - g(f(X))\|_F^2, \quad (1)$$

where $m$ is the number of samples and $\|\cdot\|_F$ is the Frobenius norm for matrices. By optimizing the above function, we can obtain an autoencoder which compresses the observed matrix $X$ as low-dimensional data $f(X)$ and outputs the decoded matrix $\hat{X} = g(f(X))$.

**Feature Selector.** Fcn. 1 investigates the self representability of autoencoder networks. We will further explore an unsupervised feature selection method by excavating redundancy in observed features based on the motivation that, if the reconstructed data $\hat{X}$ is similar to the original data $X$ after discarding some features, these features could be recognized as redundancy and discarded as shown in Fig. 1.

Denote the weight matrix connecting the input layer and the hidden layer $W^{(1)} = [\boldsymbol{w}_1, \cdots, \boldsymbol{w}_d]^T$, where the $i$-th row $\boldsymbol{w}_i$ corresponds to the $i$-th feature $\boldsymbol{x}_i$. If $\|\boldsymbol{w}_i\|_2 \approx 0$, the $i$-th feature contributes little to the representation of other features; on the other hand, if the $i$-th feature plays important role in the representation of other features, $\|\boldsymbol{w}_i\|_2$ must be significant. In order to select the most discriminative features from observed data $X$, we impose row-sparse regularization on $W^{(1)}$, *i.e.*,

$$\|W^{(1)}\|_{2,1} = \sum_i^d \sqrt{\sum_j^h (W_{ij}^{(1)})^2}. \quad (2)$$

Therefore, we reformulate Fcn. 1 as

$$\mathcal{J}(\Theta) = \frac{1}{2m}\|X - g(f(X))\|_F^2 + \alpha\|W^{(1)}\|_{2,1}, \quad (3)$$

where $\alpha$ is the trade-off parameter of the reconstruction loss and the regularization term.

In the training process of a neural network, a weight decay term is also necessary in order to avoid overfitting and promote convergence. Thus, we combine the above function and the weight decay regularization to form the following object function:

$$\mathcal{J}(\Theta) = \frac{1}{2m}\|X - g(f(X))\|_F^2 + \\ \alpha\|W^{(1)}\|_{2,1} + \frac{\beta}{2}\sum_{i=1}^{2}\|W^{(i)}\|_F^2, \quad (4)$$

where $\beta$ is the penalty parameter. By minimizing 4 over the train dataset, we can obtain an autoencoder network for selecting useful features, namely, AutoEncoder Feature Selector (AEFS).

**Nonlinearity Discussion.** As for the nonlinear property, we compare our method AEFS with a recent work related to ours, *i.e.*, regularized self-representation model (RSR) [14], which solves the following optimization function:

$$\min_W \|X - XW\|_F^2 + \lambda\|W\|_{2,1} \quad (5)$$

where $W$ is the feature weight matrix each feature. In RSR, each feature can be represented as the linear combination of its relevant features. By using $\ell_{21}$-norm to characterize the representation coefficient matrix, RSR is effective to select representative features. However, if the correlation among features is nonlinear, RSR cannot accurately excavate their relationships. In contrast, the proposed AEFS method contains both linear and nonlinear transforms (such as sigmoid and tanh activations), therefore, each feature has the possibility that can be nonlinearly represented by relevant features. Thus, even if the correlation among features is nonlinear, AEFS could still work well. Moreover, if we set $\sigma_1(X) = X$, $\sigma_2(X) = X$ and leave out the weight decay term, AEFS reduces to a linear form:

$$\min_{W^{(1)}, W^{(2)}} \frac{1}{2m} \|X - XW^{(1)}W^{(2)}\|_F^2 + \alpha \|W^{(1)}\|_{2,1}, \quad (6)$$

which is equivalent to Fcn. 5. Therefore, AEFS can be viewed as an enhanced extension of RSR.

**Table 1.** Clustering results (ACC% ± std) of different feature selection methods on benchmark datasets.

| Dataset | AllFea | LS | MCFS | UDFS | RSR | AEFS |
|---|---|---|---|---|---|---|
| Isolet | 54.0±4.6 | 51.6±3.1 | 56.5±3.1 | 45.8±3.2 | 54.3±3.4 | **58.7±3.5** |
| warpPIE10P | 28.7±3.1 | 32.9±2.8 | 38.8±4.1 | 50.4±5.2 | 35.5±2.5 | **50.7±5.3** |
| PCMAC | 50.5±0.2 | 50.8±0.2 | 50.9±0.7 | 51.6±1.0 | 51.1±0.9 | **51.7±1.1** |
| madelon | 58.2±0.5 | 58.4±0.2 | 59.1±0.3 | 58.7±0.2 | 51.3±1.1 | **61.0±0.1** |
| lung_discrete | 64.3±7.1 | 65.1±9.7 | 70.3±8.4 | 68.9±6.8 | 71.6±5.8 | **71.6±7.2** |
| Prostate_GE | 59.9±1.9 | 57.5±4.6 | 59.9±5.0 | 64.5±3.8 | 60.5±5.2 | **73.1±6.4** |
| MNIST | 46.8±2.6 | 31.4±1.7 | 50.9±2.3 | 49.0±2.7 | 29.3±0.8 | **51.8±4.8** |

## 3. OPTIMIZATION

Since the proposed method utilizes an autoencoder network for implementing the feature selection problem, whose parameters can be learned through huge samples. Therefore, we use the back-propagation strategy to optimize the autoencoder network. Firstly, the error terms of the output layer and the hidden layer are computed as follows.

$$\begin{aligned} \delta^{(o)} &= -(X - \hat{X}) \odot \sigma_2'(f(X)), \\ \delta^{(h)} &= \left((W^{(2)})^T \delta^{(o)}\right) \odot \sigma_1'(X), \end{aligned} \quad (7)$$

where $\odot$ denotes the element-wise product. The partial derivative respects to $W^{(2)}$ can be calculated as

$$\nabla_{W^{(2)}} \mathcal{J}(\Theta) = \frac{1}{m} \delta^{(o)} \hat{X}^T + \beta W^{(2)}, \quad (8)$$

and $W^{(2)}$ are optimized by gradient descent method.

However, the partial derivative of the object function respect to $W^{(1)}$ is not available at the zero point, so it can not be directly optimized by gradient descent method. Instead, we use the proximal gradient descent method [15, 16] to solve the problem. The solving process includes two steps:

$$\nabla_{W^{(1)}} \mathcal{J}^-(\Theta) = \frac{1}{m} \delta^{(h)} f(X)^T + \beta W^{(1)}. \quad (9)$$

$$\hat{W^{(1)}} = \Phi^\# \left(W^{(1)} - t\nabla_{W^{(1)}} \mathcal{J}^-(\Theta); \alpha t\right) \quad (10)$$

where $\mathcal{J}^-(\Theta)$ denotes the object function leaving out $\ell_{21}$ regularization, $t > 0$ is a step size, $\Phi^\#$ is the group soft thresholding operator and the details are described in Definition 1.

**Definition 1.** The multivariate soft thresholding operator for any vector $\boldsymbol{w} \in \mathbb{R}^d$ is $\overrightarrow{\Phi}(\boldsymbol{w}; \lambda) = \boldsymbol{w}^o \Phi(\|\boldsymbol{w}\|_2; \lambda)$ where

$$\boldsymbol{w}^o = \begin{cases} \frac{\boldsymbol{w}}{\|\boldsymbol{w}\|_2}, & \text{if } \boldsymbol{w} \neq \boldsymbol{0} \\ \boldsymbol{0}, & \text{if } \boldsymbol{w} = \boldsymbol{0} \end{cases}, \quad (11)$$

and $\Phi$ is element-wise soft thresholding operator: $\Phi(x; \lambda) = \text{sign}(x)(|x| - \lambda)_+$. Then we define the group soft thresholding operator for any matrix $W = [\boldsymbol{w}_1, \boldsymbol{w}_2, \cdots, \boldsymbol{w}_n]^T$ as

$$\Phi^\#(W; \lambda) = \begin{pmatrix} \overrightarrow{\Phi}(\boldsymbol{w}_1; \lambda)^T \\ \overrightarrow{\Phi}(\boldsymbol{w}_2; \lambda)^T \\ \vdots \\ \overrightarrow{\Phi}(\boldsymbol{w}_n; \lambda)^T \end{pmatrix}. \quad (12)$$

## 4. EXPERIMENTS

### 4.1. Datasets and Experiemntal Settings

Experiments are conducted on 7 benchmark datasets to evaluate the performance of AEFS. The datasets include one spoken letter dataset (*i.e.*, Isolet [17]), one face image dataset (*i.e.*, warpPIE10P [18]), one text dataset (*i.e.*, PCMAC [19]), one artificial dataset(*i.e.*, madelon [20]), two microarray datasets (*i.e.*, lung_discrete [21] and Prostate_GE [22]), and one handwritten digits dataset (*i.e.*, MNIST [23]). All the data is normalized before experiments.

In order to evaluate superiority of our method, we compare AEFS with the following unsupervised feature selection methods.

**AllFea**: All original features without feature selection.

**LS**: Laplacian Score [24] feature selection method which selects features that well preserve the data manifold structure.

**MCFS**: Multi-Cluster Feature Selection [25] method which selects features using spectral regression with $\ell_1$ norm regularization.

**UDFS**: Unsupervised Discriminative Feature Selection [26] method that selects features by exploiting both the discriminative information and feature correlations.

**Table 2.** Classification results (ACC%) of different feature selection methods on benchmark datasets.

| Dataset | AllFea | LS | MCFS | UDFS | RSR | AEFS |
|---|---|---|---|---|---|---|
| Isolet | 90.128 | 83.562 | **89.615** | 82.436 | 85.0 | 89.167 |
| warpPIE10P | 100.0 | 94.286 | **100.0** | 99.524 | 99.048 | **100.0** |
| PCMAC | 77.458 | 65.878 | 70.201 | 74.472 | 66.341 | **76.531** |
| madelon | 52.962 | 68.423 | 64.346 | 70.192 | 51.462 | **70.769** |
| lung_discrete | 83.562 | 85.256 | 89.041 | 89.041 | 87.671 | **90.411** |
| Prostate_GE | 80.392 | 62.745 | 81.373 | **88.235** | 79.412 | 87.255 |
| MNIST | 95.006 | 60.591 | 95.257 | 91.921 | 75.479 | **96.204** |

**RSR**: Regularized Self-Representation [14] model for feature selection which exploiting the self-representation ability of features with $\ell_{21}$ regularization.

As for parameters setting, in the methods LS, MCFS and UDFS, the size of the neighbors $k$ is fixed as 5 for all the cases. For fair comparison, the parameters in all the methods are tuned in the range of $\{0.001, 0.01, \cdots, 100, 1000\}$. In AEFS, we set the size of hidden layer in $\{128, 256, 512, 1024\}$ and the activation function $\sigma_1(X) = 1/(1 + e^{-X})$, $\sigma_2(X) = X$. For all datasets, we set the number of selected features as $\{50, 100, 150, \cdots, 300\}$ and report the best results from the optimal parameters for all the methods.

### 4.2. Clustering and Classification Experiments

Following experiment setting in [14], we conduct clustering experiments using $k$-means algorithm and classification experiments using the nearest neighbor classifier to evaluate the performance of different feature selection methods.

**Evaluation metrics.** For clustering and classification experiments, accuracy (ACC) is used to measure the performance[1].

In clustering, denote $p_i$ as the true label and $q_i$ as the clustering result of the sample $\boldsymbol{x}_i$. ACC is defined as

$$ACC = \frac{\sum_{i=1}^{m} \delta(p_i, map(q_i))}{m} \quad (13)$$

where $\delta(x, y) = 1$ if $x = y$; $\delta(x, y) = 0$ otherwise and $map(q_i)$ is the best mapping function that permutes clustering labels to match the ground truth labels, which can be gotten using the Kuhn-Munkres algorithm.

**Experimental results.** The clustering results are shown in Table 1, and the classification results are listed in Table 2. From the results, we observe that feature selection can not only reduce the dimension of features, but also greatly improve both the clustering and the classification performance. We also see that AEFS outperform other methods almost in all the cases. This benefits from the ability to capture the most import features which could represent all the features and the nonlinearity transformation inside the representation of AEFS.

---

[1]Since the result of $k$-means depends on initialization, we repeat the experiments 20 times with random initialization and report the average results with standard deviation.

### 4.3. Reconstruction Experiments

We conduct reconstruction experiments on the face dataset warpPIE10P using AEFS and RSR. The comparison results are shown in Fig. 2. The large weights of features learned by AEFS mainly distribute in the area of eyebrow, eye, nose and mouth which are important for recognition, while the weights learned by RSR is discriminative only in eye position and the eyebrow, nose and mouth are not distinct from other parts. It can be found in Fig. 2 that both AEFS and RSR can well reconstruct the raw face with much fewer features than the original, but the proposed AEFS can provide clearer reconstruction results.

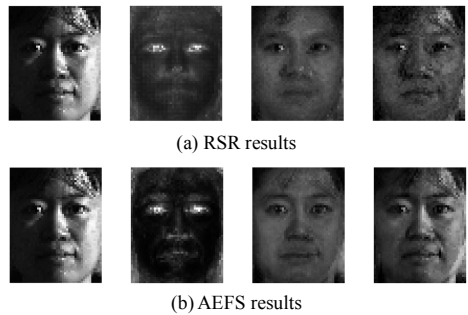

(a) RSR results

(b) AEFS results

**Fig. 2**. Face reconstruction results by RSR and AEFS, from left to right: raw face, feature weight map, reconstructed face using 300 feature, and reconstructed face using 1000 features.

## 5. CONCLUSIONS

Here we propose a novel unsupervised feature selection method which could jointly learn a self-representation autoencoder model and the importance weights of each feature. The autoencoder nonlinearly represent each feature using all the features with different weights. By minimizing the reconstruction error and the group sparsity regularization simultaneously, we obtain a subset of observed features which can preserve intrinsic information of the original data. Experimental results on several benchmark datasets validate the superiority of our methods over other unsupervised feature selection methods.